\documentclass[10pt,twocolumn,letterpaper]{article}

\usepackage{iccv}
\usepackage{times}
\usepackage{epsfig}
\usepackage{graphicx}
\usepackage{amsmath}
\usepackage{amssymb}
\usepackage{booktabs}
\usepackage{multirow}
\usepackage{lipsum}

\usepackage[pagebackref=true,breaklinks=true,letterpaper=true,colorlinks,bookmarks=false]{hyperref}

\iccvfinalcopy %

\def\iccvPaperID{37} %

\newcommand*{\affaddr}[1]{#1} 
\newcommand*{\affmark}[1][*]{\textsuperscript{#1}}

\newcommand\blfootnote[1]{%
  \begingroup
  \renewcommand\thefootnote{}\footnote{#1}%
  \addtocounter{footnote}{-1}%
  \endgroup
}

\ificcvfinal\fi

\begin{document}

\title{On Adversarial Robustness: A Neural Architecture Search perspective}
\author{%
Chaitanya Devaguptapu\affmark[1]\hspace{0.5em} Devansh Agarwal\affmark[1]\thanks{Equal contribution} \hspace{0.5em} Gaurav Mittal\affmark[2]\footnotemark[1]  \hspace{0.5em} 
Pulkit Gopalani \affmark[3] \\ \hspace{0.5em} Vineeth N Balasubramanian\affmark[1] \\
\affaddr{\affmark[1]Indian Institute of Technology, Hyderabad, India}
\\ \affaddr{\affmark[2]Microsoft} \\
\affaddr{\affmark[3]Indian Institute of Technology, Kanpur, India} \\
}

\clearpage\maketitle
\thispagestyle{empty}

 \blfootnote{Corresponding author: cs19mtech11025@iith.ac.in}

\begin{abstract}
Adversarial robustness of deep learning models has gained much traction in the last few years. Various attacks and defenses are proposed to improve the adversarial robustness of modern-day deep learning architectures. 
While all these approaches help improve the robustness, one promising direction for improving adversarial robustness is unexplored, \ie, the complex topology of the neural network architecture. 
In this work, we address the following question: \textit{``Can the complex topology of a neural network give adversarial robustness without any form of adversarial training?''}. We answer this empirically by experimenting with different hand-crafted and NAS-based architectures. 
Our findings show that, for small-scale attacks, NAS-based architectures are more robust for small-scale datasets and simple tasks than hand-crafted architectures. 
However, as the size of the dataset or the complexity of task increases, hand-crafted architectures are more robust than NAS-based architectures. 
Our work is the first large-scale study to understand adversarial robustness purely from an architectural perspective. 
Our study shows that random sampling in the search space of DARTS (a popular NAS method) with simple ensembling can improve the robustness to PGD attack by nearly~12\%. 
We show that NAS, which is popular for achieving SoTA accuracy, can provide adversarial accuracy as a free add-on without any form of adversarial training.
Our results show that leveraging the search space of NAS methods with methods like ensembles can be an excellent way to achieve adversarial robustness without any form of adversarial training.
We also introduce a metric that can be used to calculate the trade-off between clean accuracy and adversarial robustness. Code and pre-trained models will be made available at \url{https://github.com/tdchaitanya/nas-robustness}

\end{abstract}

\section{Introduction} \label{intro}

\begin{figure}[]

   \includegraphics[width=\hsize, scale=0.5]{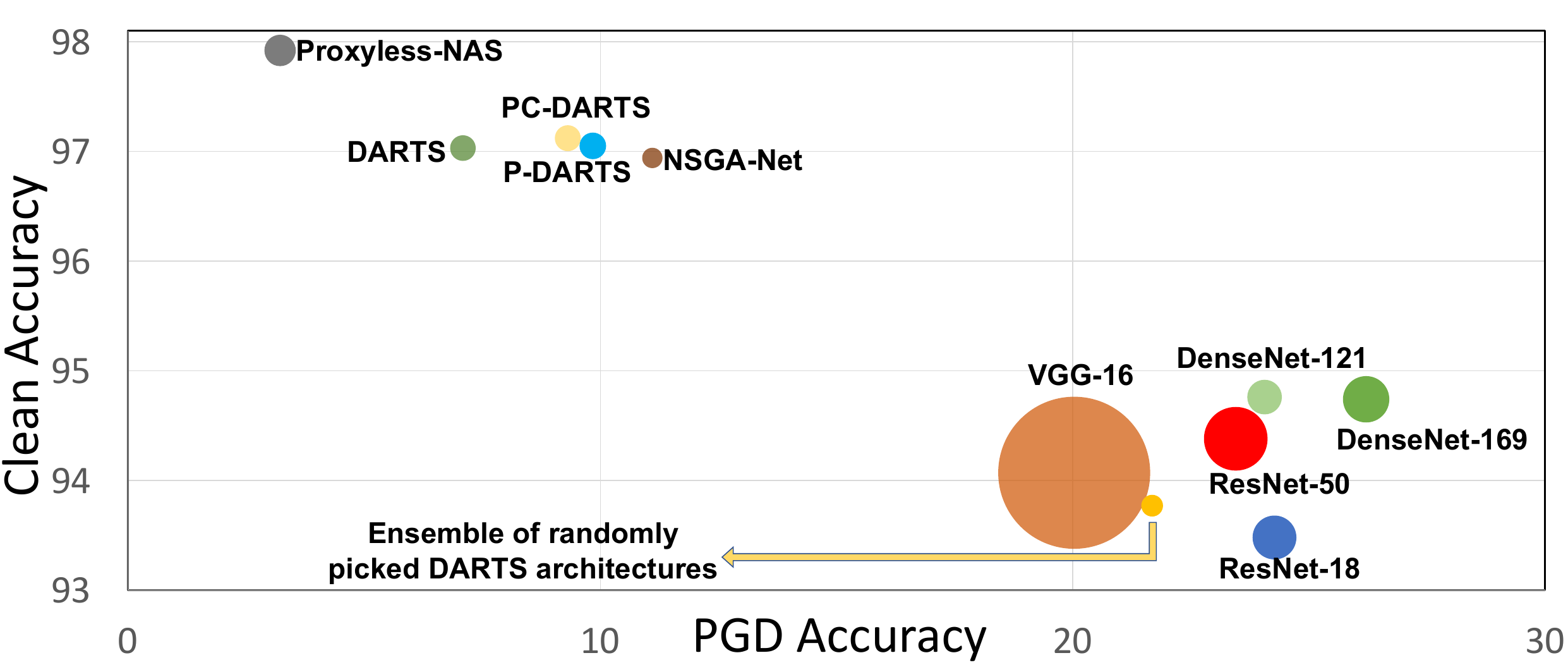}
   \caption{\textbf{NAS based architectures are \textit{slightly} better than hand-crafted architectures in terms of test accuracy. But hand-crafted architectures are \textit{significantly} more robust to PGD attacks than NAS architectures}. Qualitative comparison of test-set accuracy and PGD accuracy of NAS and hand-crafted architectures on CIFAR-10 dataset. Bubble size represents the number of parameters.}
   \vspace{-1.5em}

\label{fig:teaser}
\end{figure}
The choice of neural network architecture and its complex topology play a crucial role in improving the performance of several deep learning based applications. 
However, in most cases, these architectures are typically designed by experts in an ad-hoc, trial-and-error fashion. 
Early efforts on Neural Architecture Search~(NAS)~\cite{Zophetal2016} alleviate the pain of hand-designing these architectures by partially automating the process of finding the right topology that can result in best-performing architectures. 
Since the work by \cite{Zophetal2016}, there has been much interest in this space. Many researchers have come up with unique approaches \cite{yanetal2019, chenetal2019, phametal2018} to improve the performance besides decreasing the computational cost. 
Current SoTA (state-of-the-art) on image classification and object detection \cite{Tanetal20191,tanetal2019} are developed using NAS, which shows how important a role NAS plays in solving standard learning tasks, especially in computer vision.

Adversarial robustness is defined as the accuracy of a model when adversarial examples (images perturbed with some imperceptible noise) are provided as input. Adversarial examples have the potential to be dangerous. \cite{Practical-black-box-attacks} discusses an example where attackers could target autonomous vehicles by using stickers or paint to create an adversarial stop sign that the vehicle could interpret as a yield or other sign. One commonly used technique to improve the adversarial robustness of neural networks is adversarial training, but in most of the cases adversarial training decreases the accuracy on clean (un-perturbed) samples \cite{pgd, 2019arXiv190108573Z}. So, it is essential to develop architectures that are inherently robust without any form of adversarial training. This forms the primary motivation of our work,  \textit{Can the complex topology of a neural network architecture provide adversarial robustness without any form of adversarial training?}

In an attempt to understand adversarial robustness purely from an architectural perspective, we seek to answer the following questions, 

\begin{itemize}
    \item In the absence of adversarial training, how do NAS-based architectures compare with hand-crafted architectures (like ResNets~\cite{heetal2015}, DenseNets~\cite{huangetal2016}, \etc) in terms of adversarial robustness?
    \item Does an increase in the number of parameters of the architecture help improve robustness?
    \item Where does the source of adversarial vulnerability lie for NAS?  Is it in the search space or in the way the current methods are performing the search?
\end{itemize}

To the best of our knowledge, our work is the first attempt at understanding adversarial robustness purely from an architectural perspective. We show that the complex topology of neural network architectures can be leveraged to achieve robustness without adversarial training.  Additionally, we introduce two simple metrics, \textit{Harmonic Robustness Score (HRS)} and \textit{Per-parameter HRS (PP-HRS)} that combine: (1) the total number of parameters in a model; and (2) accuracy on both clean and perturbed samples, to convey how robust and deployment-ready a given model is when no adversarial training is performed. 

We examine the adversarial robustness of different hand-crafted and NAS-based architectures in a wide range of scenarios and find that for large-scale datasets, complex tasks and stronger attacks (like PGD~\cite{pgd}), traditional hand-crafted architectures like ResNets and DenseNets are more robust than NAS-based architectures (Figure \ref{fig:teaser}). This suggests that the adversarial robustness of a model depends significantly on network topology. Results of our study can be used to design network architectures that can give adversarial robustness with no additional adversarial training along with SoTA performance on unperturbed samples.

\section{Related Work}\label{relatedWork}

\paragraph{Adversarial Attacks and Robustness:} Adversarial examples, in general, refer to samples that are imperceptible to the human eye but can fool a deep classifier to predict a non-true class with high confidence. Adversarial examples can result in degraded performance even in the presence of perturbations too subtle to be perceived by humans.

Existing adversarial attacks can be broadly classified into white-box and black-box attacks. The difference between these lies in the knowledge of the adversaries. In white-box attacks, the adversaries have the full knowledge of the target model, including the model architecture and parameters.  In a black-box setting, the adversaries knowledge is very limited and may not know details about the model.

In the frameworks of these threat models, several effective adversarial attacks have been proposed over the years such as L-BFGS~\cite{szegedyetal2013}, FGSM~\cite{fgsm}, BIM~\cite{bim}, C\&W attacks~\cite{cnwattack} JSMA~\cite{jsma}, Deep-Fool~\cite{deepfool}, R-FGSM~\cite{rfgsm},  StepLL~\cite{stepll}, PGD~\cite{pgd} and most recently SparseFool~\cite{Modas_2019_CVPR}, F-FGSM~\cite{f-fgsm} and AutoPGD~\cite{autopgd}. For more information on adversarial attacks and defenses, please see \cite{chakrborty2018, REN2020346}. White-box is a stronger setting where attackers can access the model parameters and architecture. It is also closely related to the network topology aspect of our study. So we mainly focus on the white-box setting in our work and present some results on black-box attacks in the Appendix.

One popular way to improve the adversarial robustness of deep learning models is adversarial training (AT) \cite{goodfellowetal2014}. The basic idea of AT is to create and incorporate adversarial samples during training. A critical downside of AT is that it is time-consuming\cite{NIPS2019_8597}. In addition to the gradient computation needed to update the network parameters, each stochastic gradient descent (SGD) iteration requires multiple gradients computations to produce adversarial images.

\textbf{Neural Architecture Search~(NAS)} automates the design of neural network architectures for a given task. Over the years, several approaches have emerged to search architectures using methods ranging from Reinforcement Learning~(RL)~\cite{Zophetal2016}, Neuro-evolutionary approaches~\cite{real2019regularized}, Sequential Decision Processes~\cite{liu2018progressive}, One-shot methods~\cite{phametal2018} and fully differentiable Gradient-based methods~\cite{Darts2018}. While most of these algorithms attempt to search a cell architecture (micro search) due to the computational cost involved and repeat the cell a fixed number of times, few recent approaches have also demonstrated searching the full architecture (macro search).

Most of the early approaches are based on RL and neuro-evolutionary algorithms, making the search process computationally intensive. Recently these have been replaced by one-shot fully-differentiable gradient-based NAS methods, such as DARTS~\cite{Darts2018}, which are orders of magnitude faster than non-differentiable techniques and have gained much traction recently.  P-DARTS~\cite{pdarts} bridges the gap between search and evaluation by progressively increasing search depth. Partially-Connected DARTS~\cite{pcdarts}, a SoTA approach in NAS, significantly improves the efficiency of one-shot NAS by sampling parts of the super-network and adding edge normalization to reduce redundancy and uncertainty in search. DenseNAS~\cite{densenas}, a more recent method, attempts to improve search space design by further searching block counts and block widths in a densely connected search space. Despite a plethora of these methods and their applications, there has been minimal effort to understand the adversarial robustness of final learned architectures.

\textbf{Adversarial Robustness of Architectures: }
\cite{pgd} is one of the early papers to talk about adversarial robustness of network architectures. It shows that when training with unperturbed samples, increasing the capacity of the network in terms of width, depth, and the number of parameters can alone help improve the robustness for datasets like MNIST and CIFAR-10. Recently, \cite{xieetal2019} echoes this observation by showing the depth of the network helps to improve the adversarial robustness during adversarial training. Both \cite{pgd, xieetal2019} talk about robustness mainly in the context of adversarial training. However, our results show that when no adversarial training is performed, increasing parameters alone only helps to a certain point and beyond that, it reduces the adversarial robustness of the model.

Very recently, there have been limited efforts to improve adversarial robustness using architecture search~\cite{guo2019meets, vargas2019evolving}. \cite{guo2019meets} proposes a robust architecture search framework by leveraging one-shot NAS. However, the proposed method adversarially trains the entire NAS search space before starting the search process, making it harder to assess the contribution of just the architecture to the adversarial robustness. \cite{vargas2019evolving} uses black-box attacks to generate a fixed set of adversarial examples on CIFAR-10 and uses these examples to search for a robust architecture using NAS. The experimental setting is constrained and does not reflect the true robustness of the model as the adversarial examples are fixed a priori. No study is done on white-box attacks. Both \cite{guo2019meets} and \cite{vargas2019evolving} do not make any comparisons with existing NAS methods (which, as per our study, are already robust to an extent).

In this work, we mainly focus on evaluating the robustness of SoTA NAS methods on white-box attacks across datasets of different sizes, including large-scale datasets such as ImageNet~\cite{imagenet_cvpr09} and compare them with hand-crafted models like ResNets and DenseNets.  As a part of our study, we introduce metrics that can be used to estimate the trade-off between clean accuracy and adversarial robustness when comparing architectures within and across different families.

\section{Robustness of NAS models: A study}\label{experiments}

We carefully design our experimental setting to answer the questions stated in Section~\ref{intro}. We begin by describing the design of our experiments, providing details about datasets, models, attacks, and metrics. \\

\noindent \textbf{Datasets:} Since we want to compare the robustness of architectures across different dataset scales and complexities, we choose four different image classification datasets. In addition to the standard CIFAR-10 \cite{cifar10} dataset, which consists of 60K images of $32 \times 32$ resolution, we also choose CIFAR-100 \cite{cifar100} to test if the same robustness trends hold when the labels turn from coarse to more fine-grained and the number of classes increase by a factor of 10. To study the robustness trend for tougher tasks like fine-grained image classification where the classes are semantically and perceptually more similar, we choose Flowers-102 dataset~\cite{flowers102}, which consists of 8189 flowers images split across 102 categories with number of images in each category being between 40 and 258. 
Since most real-world applications deal with large-scale datasets, we also test robustness on ImageNet~\cite{imagenet_cvpr09} dataset, consisting of $\sim$1.3M images from 1000 classes. This makes our study more complete when compared to earlier works. 

\noindent \textbf{Architectures:} We select most commonly used NAS methods including DARTS~\cite{Darts2018}, P-DARTS~\cite{pdarts}, ProxylessNAS~\cite{cai2018proxylessnas}, NSGA-Net~\cite{nsganet}, along with recent methods like PC-DARTS~\cite{pcdarts} and DenseNAS~\cite{densenas}.  We evaluate five well-known handcrafted architectures and at least four NAS architectures on each dataset mentioned above for a fair comparison. For all experiments, we either use pre-trained models made available by the respective authors or train the models from scratch until we obtain the performance reported in the respective papers. For the results on Flowers-102 dataset, we explicitly search for an architecture using the code provided by \cite{Yang2020NAS}. The results for NSGA-Net are only available for CIFAR-10/100 because its implementation does not support Imagenet. Similarly, the implementation of DenseNAS does not support CIFAR-10/100, so the results are shown only for ImageNet. ProxylessNAS provides pre-trained models for only CIFAR-10 and ImageNet, so we show results only for these two datasets.

\begin{figure*}
\begin{center}
\includegraphics[width=15cm, height=5.7cm]{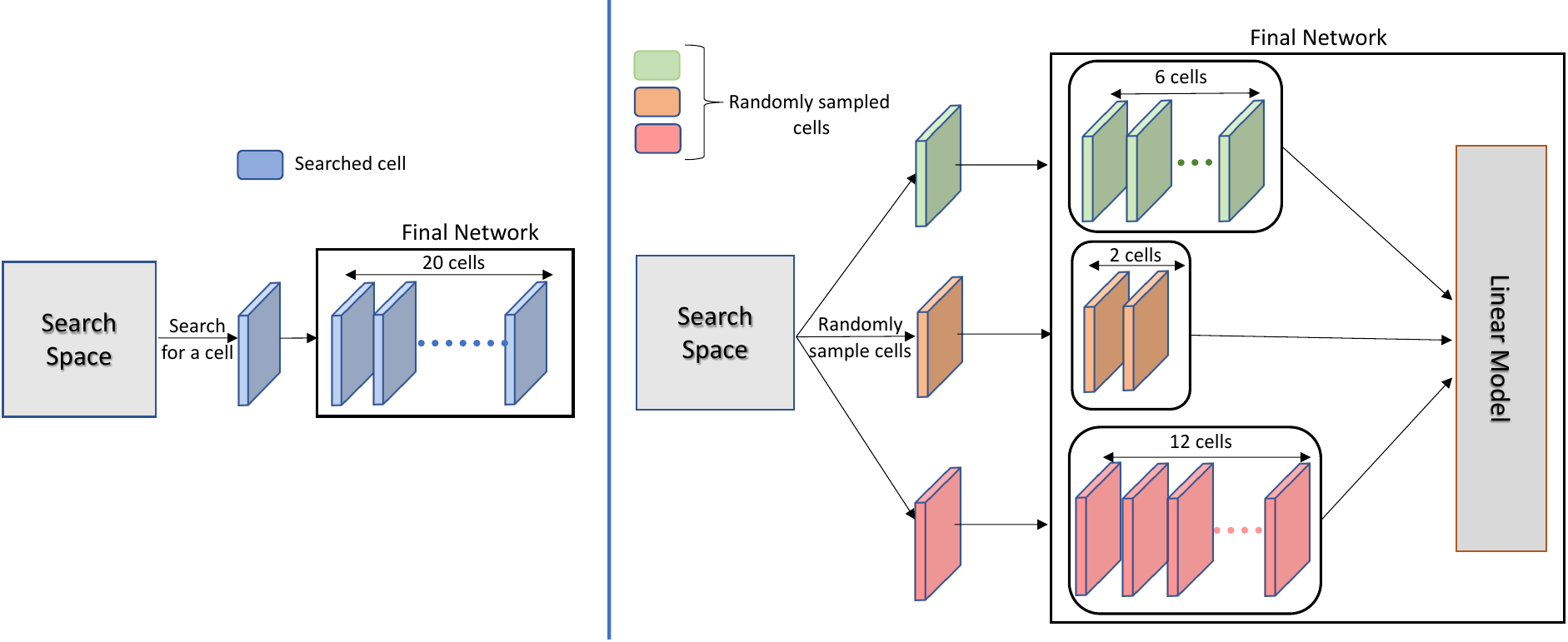}
\end{center}
   \caption{\textit{Left: } Standard procedure for building architectures from DARTS search space; \textit{Right: } Procedure for building ensembles using DARTS search space. 12, 6, 2 can be replaced with any values that sum to 20.} \label{robust_comparison}
\vspace{-0.5cm}
\label{comparison}
\end{figure*}

\noindent \textbf{Ensemble of Architectures:} When compared with single architecture, an ensemble of architectures are known to be adversarially more robust \cite{pmlr-v97-pang19a}. To understand the effectiveness of ensembling, in Section~\ref{question4}, we random sample cells from the DARTS search space using the code provided by \cite{Yang2020NAS}; and stack these cells to create small architectures, since this is randomly sampling, the search cost associated with building these architectures is zero. After sampling, we follow the standard DARTS training protocol to train these architectures. In general, for the CIFAR-10 dataset, DARTS architectures having 20 cells are trained for 600 epochs. Following this, the number of epochs for training each network in the ensemble is determined based on the number of cells in that network. Effectively the ensemble as a whole is trained for 600 epochs to ensure we make a fair comparison with existing approaches. After training each of these networks separately, we train a simple linear model to combine the individual model outputs. This linear model is trained only for two epochs. The difference between standard DARTS and ensembling by sampling from DARTS search space is visually shown in Figure \ref{comparison}. More details on the structure of the linear model are discussed in Section \ref{question4}

\noindent \textbf{Adversarial Attacks:} \label{advattacks} For adversarial robustness, we test against the standard attacks like FGSM~\cite{fgsm}, PGD~\cite{pgd} and also report results on recently introduced F-FGSM~\cite{f-fgsm} and AutoPGD~\cite{autopgd}. For all these attacks, we use a perturbation value of $8/255 (0.03)$ which denotes the maximum noise added to each pixel in the input image as perturbation. The step size is $2/255~(0.007)$ with the attack iterations set as 10. Moreover, we run the AutoPGD attack with 10 random restarts. All these parameter choices are standard and widely used in the community~\cite{2020arXiv201000467P,2020arXiv200806081F, f-fgsm}. Architectures are trained using standard training protocols, and no adversarial training is performed. We use the library provided by \cite{kim2020torchattacks} for all the adversarial attacks in our experiments.

\noindent \textbf{Metrics:} We use Clean Accuracy and Adversarial Accuracy as our performance metrics. Clean accuracy refers to the accuracy on the unperturbed test set as provided in the dataset. For each attack, we measure Adversarial accuracy by perturbing the test set examples using various attacks in the methods listed in the above section~(FGSM, F-FGSM, PGD and AutoPGD). 

 One of the main problems with adversarially trained models is that their clean accuracy is usually less than standard non-adversarially trained models. Adversarial vulnerability is a side-effect of overfitting to the training set~\cite{2020arXiv200704028S}. While this overfitting gives good performance on the clean test set, it makes the model vulnerable to adversarial examples. If the accuracy of a model on clean samples is not good, it is not useful when deployed in situations where unperturbed samples are more frequent. On the other hand, if the model has SoTA performance on a clean test-set, it becomes vulnerable to adversarial examples. There is no well-defined metric to capture this trade-off between clean and adversarial accuracy.

To this end, we introduce a metric, called~\textbf{\textit{Harmonic Robustness Score ($\text{HRS}$)}}, that is defined as the harmonic mean of the clean and adversarial accuracy of a given model.  $\text{HRS}$ captures the balance of a model's performance to unperturbed inputs and robustness to an adversarial attack
. Consider a model with clean accuracy C and Adversarial accuracy A (both in percentage), $\text{HRS}$ for that model is calculated as follows:
 \begin{equation}
    \text{HRS} = 2\cdot\frac{\text{C}\cdot\text{A}}{\text{C}+\text{A}}
\end{equation}
The harmonic mean is better at reflecting extreme differences in input values, compared to Arithmetic mean. Therefore, if one of the clean or adversarial accuracy is \textit{very} low, then the harmonic mean of C and A would be more reflective of the same.

A weighted version of $\text{HRS}$, the $\text{HRS}_{\beta}$ score, can also be used. This is a measure of the model's performance to clean as well as perturbed images, weighted according to what the end-user prefers -- clean accuracy, or the adversarial accuracy. For a given use case, one might be preferred more to the other, and hence this metric can be used accordingly. 
$\text{HRS}_{\beta}$ is given by,
 \begin{equation}
    \text{HRS}_{\beta} = (\beta^2 + 1)\cdot \frac{\text{C}\cdot\text{A}}{\beta^2 \text{C} + \text{A}}
\end{equation}
where $\beta$ can be interpreted as the importance of adversarial accuracy over clean accuracy. Since we do not have any particular preference to adversarial accuracy over clean accuracy (or vice-versa), we use $\beta=1$ for reporting HRS values. The adversarial accuracy for all models is measured on perturbed inputs obtained using the PGD \cite{pgd} attack. (Choice of PGD is arbitrary, and can be replaced with any other attack).

When comparing performances of architectures belonging to the same family, number of parameters play an important role. A huge parameter difference can easily improve clean and adversarial accuracy, but this comes with huge training and inference time. 
So we further define per-parameter harmonic robustness score (PP-HRS) to measure the clear accuracy verses adversarial robustness trade-off within a family of architectures. PP-HRS compares the parameters of the model with the parameters in the baseline model of that family. In a family of architectures ( $\mathcal{F}$ ), consider a baseline model $m_b$ having $p_b$ number of parameters, now for a model $m_i \in \mathcal{F}$ with $p_i$ number of parameters, PP-HRS is calculated as follows,
\begin{equation}
    \text{PP-HRS} = \text{HRS} * \frac{p_b}{f(p_i)}
\end{equation}
where the function $f(p_i)$ can be defined as per requirement. We use $f(p_i) = p_i$ as our function of choice for PP-HRS. The main motivation behind this choice is to consider the improvement of accuracy with number of parameters. For example, in the EfficientNet~\cite{Tanetal20191} family, we can compute the average increase in accuracy per unit increase in the number of parameters, for the purpose of comparison.

\section{Analysis and Results} \label{analysis and Results}

In this section, we compare and contrast the robustness of different architectures in a wide-range of scenarios and answer questions listed in Section~\ref{intro}.

\begin{figure*}
\begin{center}
\includegraphics[scale=0.23]{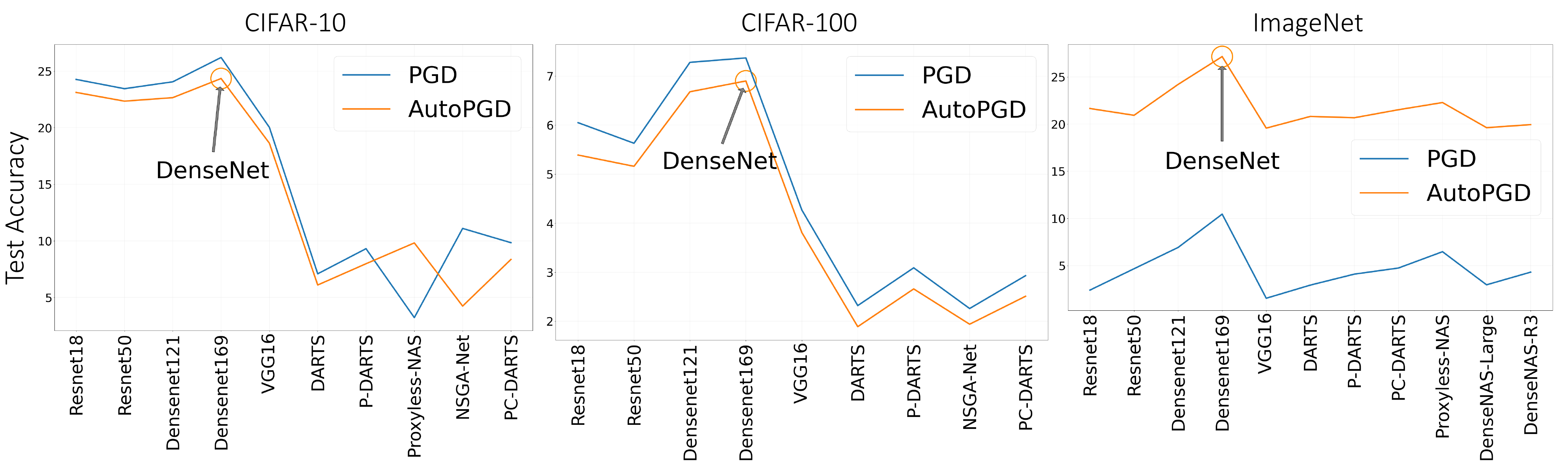}
\end{center}
   \caption{\textbf{DenseNets are always more robust}. Qualitative comparison of accuracy of different models for PGD and AutoPGD attacks on CIFAR-10, CIFAR-100 and ImageNet datasets} \label{robust_comparison}
\label{robust_comparison}
\end{figure*}

\subsection{How do NAS-based models compare with hand-crafted models in terms of architectural robustness?} \label{question1}

The HRS and robustness of different hand-crafted and NAS-based architectures on CIFAR-10, CIFAR-100, ImageNet and Flowers-102 datasets are shown in Tables \ref{c10}, \ref{c100}, \ref{imagenet}, \ref{f102} respectively. 

In the case of CIFAR-10 and CIFAR-100, NAS-based architectures outperform hand-crafted architectures in terms of architectural robustness for attacks like FGSM and F-FGSM by a significant margin. However, for stronger and most commonly used attacks like PGD and AutoPGD, NAS-based architectures fail significantly compared to hand-crafted models. In terms of HRS, for CIFAR-10 dataset, the difference in the best-performing NAS and hand-crafted models is 21\%.

\begin{table}[h]
\scalebox{0.72}{\begin{tabular}{lcccccc}
\toprule
Model                                                     & Clean \%       & FGSM     & F-FGSM   & PGD    &AutoPGD  &HRS      \\ \toprule
ResNet-18                                                 & 93.48          & 52.43          & 48.33    & 24.27      & 23.13  &38.53   \\
ResNet-50                                                 & 94.38          & 50.05          & 45.78          & 23.45    & 22.35     &37.57 \\
DenseNet-121                                              & 94.76          & 50.94          & 47.14          & 24.06     & 22.66    &38.38  \\
DenseNet-169                                              & 94.74          & 53.53          & 49.47          & \textbf{26.21}  & \textbf{24.35} &\textbf{41.06}\\
VGG16 BN                                                  & 94.07          & 52.42          & 46.16          & 20.03       & 18.63    & 33.03 \\ \midrule
DARTS \cite{Darts2018}                   & 97.03          & 58.53          & 45.03          & 7.09         & 6.10  &13.21\\
PDARTS \cite{pdarts}                     & \textbf{97.12} & 58.67          & 47.62          & 9.31         & 7.98   & 16.99\\
NSGA Net \cite{nsganet}                  & 96.94          & \textbf{66.08} & 56.16          & 11.1      & 9.82     &19.92 \\
Proxyless-NAS \cite{cai2018proxylessnas} & 97.92          & 51.73          & \textbf{58.38} & 3.22       & 4.24    &6.23\\
PC-DARTS \cite{pcdarts}                  & 97.05          & 60.55          & 48.65          & 9.84         & 8.36   &17.87\\ \bottomrule
\end{tabular}}
\vspace{0.05cm}\caption{\textbf{For simple attacks, NAS based architectures are robust, but for strong attacks hand-crafted architectures are better}. Quantitative comparison of clean accuracy and adversarial robustness on \textbf{CIFAR-10} dataset (Top-1 Accuracy)}
\label{c10}
\vspace{-0.2cm}
\end{table}

\begin{table}[h]
\scalebox{0.72}{\begin{tabular}{lcccccc}
\toprule
Model                      & Clean \%       & FGSM           & F-FGSM        & PGD    &AutoPGD      &HRS  \\ \toprule
ResNet-18                  & 63.87          & 17.08          & 17.12         & 6.05     &5.39    &11.05 \\
ResNet-50                  & 73.09          & 19             & 18.12         & 5.63     &5.16     & 10.45 \\
DenseNet-121               & 78.71          & 22.9           & 22.22         & 7.28     &6.68   & 13.33 \\
DenseNet-169               & 82.44          & 22.73          & 21.66         & \textbf{7.37} &\textbf{6.90} & \textbf{13.53}\\
VGG16 BN                   & 72.05          & 17.09          & 15.15         & 4.27         &3.81  & 8.06\\ \midrule
DARTS  \cite{Darts2018}   & 82.43          & 24.91          & 16.34         & 2.32        &1.89  & 4.51 \\
PDARTS   \cite{pdarts}    & 83.07          & 27.69          & 20.23         & 3.09        &2.66  & 5.96\\
NSGA Net   \cite{nsganet} & \textbf{85.44} & \textbf{34.93} & \textbf{24.1} & 2.26     &1.94 & 4.40    \\
PC-DARTS  \cite{pcdarts}  & 81.83          & 26.22          & 18.35         & 2.93     &2.51 & 5.66    \\ \bottomrule
\end{tabular}}
\vspace{0.05cm}\caption{\textbf{For simple attacks, NAS based architectures are robust, but for strong attacks hand-crafted architectures are better}. Quantitative comparison of clean accuracy and adversarial robustness on \textbf{CIFAR-100} dataset (Top-1 Accuracy)}
\label{c100}
\vspace{-0.5cm}
\end{table}

This trend seen in CIFAR-10/100 for attacks like FGSM and F-FGSM do not hold for large-scale datasets like ImageNet and relatively complex tasks like fine-grained classification. In the case of Imagenet, handcrafted models are more robust than NAS-based architectures for all the attacks. Similarly, for the task of fine-grained classification on Flowers-102 dataset, handcrafted models like DenseNet-169 and VGG-16 beat NAS based architectures by a significant margin. Even in terms of clean accuracy, for which the NAS-based models are generally known to be better than handcrafted models, NAS architectures fail by a margin of $\sim$1.5\% for the Flowers-102 dataset.

This trend of robustness for all four datasets is clearly shown in Figure \ref{robust_comparison}. As the dataset size or the task complexity increases, hand-crafted models start to be better for all the three adversarial attacks. For stronger attacks like PGD, handcrafted models are more robust when compared to NAS-based architectures at any given dataset scale. While NAS-based architectures achieve SoTA clean accuracy in general, the robustness of these architectures is very erratic.

\begin{table}[h]
\scalebox{0.72}{\begin{tabular}{lcccccc}
\toprule
\multicolumn{1}{l}{Model} & Clean \% & FGSM           & F-FGSM         & PGD   &AutoPGD   &HRS       \\ \toprule
ResNet18                  & 89.08    & 32.75          & 18.03          & 2.41     &21.65 & 4.70     \\
ResNet50                  & 92.86    & 46.28          & 26.22          & 4.68     &20.93  & 8.90   \\
DenseNet121               & 91.97    & 56.20          & 38.11          & 6.932     &24.20  & 12.89   \\
DenseNet169               & 92.81    & \textbf{61.89} & \textbf{44.22} & \textbf{10.46} &\textbf{27.15} & \textbf{18.80}\\
VGG16                     & 91.52    & 33.34          & 13.54          & 1.55      &19.57 & 3.05    \\ \midrule
DARTS                     & 91.26    & 54.41          & 31.18          & 2.94     &20.81  & 5.70    \\
P-DARTS                   & 92.61    & 55.53          & 33.87          & 4.11     &20.67  & 7.86    \\
PC-DARTS                  & 92.49    & 58.90          & 37.86          & 4.75      &21.35 & 9.04    \\
Proxyless-NAS             & 92.54    & 59.56          & 39.69          & 6.48          &22.28   & 12.11   \\
DenseNAS-Large            & 92.80    & 47.91          & 27.25          & 2.97         &19.62  & 5.76 \\
DenseNAS-R3               & \textbf{93.81}    & 54.99          & 32.11          & 4.32 &19.94 & 8.25         \\ \bottomrule
\end{tabular}}
\vspace{0.05cm}\caption{\textbf{As the scale of the problem increases, hand-crafted architectures are more robust than NAS based architectures}. Quantitative comparison of clean accuracy and adversarial robustness on \textbf{ImageNet} dataset (Top-5 Accuracy)}
\label{imagenet}
\vspace{-0.4cm}
\end{table}

\begin{figure*}
\begin{center}
\includegraphics[width=16cm, height=6.5cm]{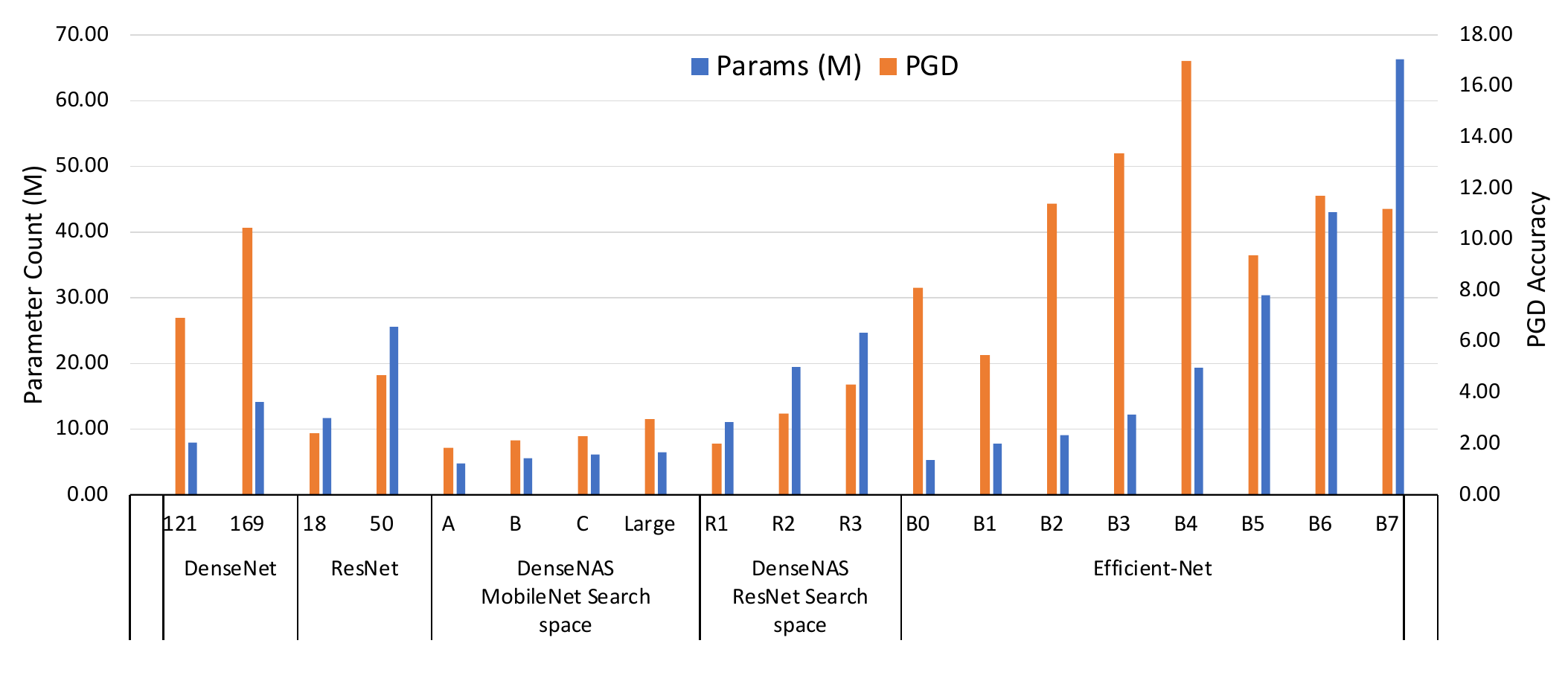}
\end{center}
   \caption{Comparison of PGD accuracy and Parameter count across different family of architectures} \label{robust_comparison}
\vspace{-0.5cm}
\label{paramcountfig}
\end{figure*}

\begin{table}[]
\scalebox{0.72}{\begin{tabular}{lcccccc}
\toprule
Model                     & Clean \% & FGSM & F-FGSM & PGD &AutoPGD & HRS  \\ \toprule
ResNet-18                 & 95.48                        & 54.33                    & 51.16                   & 11.23     &10.38      & 20.10           \\
ResNet-50                 & 97.31                        & 53.97                    & 52.38                      & 11.36    &10.01    & 20.34           \\
DenseNet-121              & 97.19                        & 67.4                     & 58.61                      & 16        &13.80   & 27.48           \\
DenseNet-169              & \textbf{97.44}               & 69.11                    & 62.76                      & 18.56    &16.48    & 31.18           \\
VGG16 BN                  & 95.24                        & \textbf{72.16}           & \textbf{66.06}             & \textbf{27.59} &\textbf{26.74 } & \textbf{42.78}       \\ \midrule
DARTS  \cite{Darts2018}  & 95.97                        & 64.47                    & 59.95                      & 19.29    &18.19     & 32.12          \\
PDARTS   \cite{pdarts}   & 95.12                        & 55.31                    & 51.16                      & 9.52     &8.55     & 17.31          \\
NSGA Net  \cite{nsganet} & 92.55                        & 40.05                    & 33.58                      & 2.69      &2.08    & 5.23          \\
PC-DARTS  \cite{pcdarts} & 94.02                        & 54.7                     & 45.3                       & 6.84    &6.23  & 12.75              \\ \bottomrule
\end{tabular}}
\vspace{0.05cm}\caption{\textbf{For difficult tasks like fine-grained classification, hand-crafted models with more parameters are robust}. Quantitative comparison of clean accuracy and adversarial robustness on Flowers-102 dataset (Top-1 Accuracy)}
\label{f102}
\vspace{-0.5cm}
\end{table}

In summary, as the dataset size~(in terms of both samples and number of classes) or the complexity of the task increases, NAS-based architectures are more vulnerable to adversarial attacks than hand-crafted models when no explicit adversarial training is performed.

\subsection{Does an increase in the number of parameters of architecture help improve robustness?} \label{question2}

\cite{dongetal2018} and \cite{pgd} observed that within the same family of architectures, increasing the number of network parameters helps improve robustness. We therefore hypothesize that increasing model capacity benefits network robustness. To study this claim, we compare the robustness of five families of architectures on the ImageNet dataset with respect to the parameter count. For comparing the trends, we use PGD accuracy along with the Per-parameter Harmonic Robustness Score (PP-HRS). The five different families of architectures we considered for this study are mentioned below.

First, we choose all the eight different variants of the EfficientNet family~\cite{Tanetal20191}. EfficientNet is a family of models that are developed by taking a NAS-based base model and scaling its width, depth and input image resolution proportionately using a set of compound-scaling coefficients which are searched via extensive grid search. We also study a recent family of SoTA NAS-based models called DenseNAS~\cite{densenas}. DenseNAS architectures are developed using two different search spaces. DenseNAS-A/B/C and Large are developed using a MobileNetV2-based search space, and DenseNAS-R1, R2, R3 are developed using a ResNet-based search space. These networks are listed in the increasing order of their parameters. Lastly, to also understand the trend in hand-crafted models, we study the robustness of standard DenseNet and ResNet models. 
 
\begin{table}[]
\scalebox{0.70}{\begin{tabular}{@{}l|lccccc@{}}
\toprule
                              Family & Variant & Params (M)  & Clean \%   & PGD    &AutoPGD     & PP-HRS   \\ \toprule
\multirow{8}{*}{Efficient-Net} & B0                 & 5.29        & 91.36   & 8.11   &25.61  & \textbf{14.90}      \\
                               & B1                 & 7.79      & 88.89     & 5.47    &24.90   & 7.00    \\
                               & B2                 & 9.11      & 92.77     & 11.40   &25.77    & 11.79   \\
                               & B3                 & 12.23      & \textbf{93.04}    & 13.37  &27.15   & 10.11     \\
                               & B4                 & 19.34      & 92.73    & \textbf{16.99} &\textbf{29.64} & 7.86 \\
                               & B5                 & 30.39      & 90.95    & 9.37   &25.28    & 2.96    \\
                               & B6                 & 43.04      & 91.86    & 11.71   &26.17  & 2.55     \\
                               & B7                 & \textbf{66.35} & 91.57 & 11.20  &27.82  & 1.59      \\ \midrule
\multirow{8}{*}{DenseNAS}      & A                  & 4.77          & 90.94 & 1.84   &\textbf{19.67}   & 3.61     \\
                               & B                  & 5.58          & 91.89 & 2.13 &19.37 & 3.56         \\
                               & C                  & 6.13          & 92.31 & 2.29    &19.22    & 3.48   \\
                               & Large              & \textbf{6.48}  & \textbf{92.80} & \textbf{2.97}  &19.62 & \textbf{4.24} \\ \cmidrule(l){2-7} 
                               & R1                 & 11.09         & 91.33 & 2.01    &19.77  & \textbf{3.93}     \\
                               & R2                 & 19.47     & 92.47     & 3.19    &19.60      & 3.51 \\
                               & R3                 & \textbf{24.66} & \textbf{93.81} & \textbf{4.32} &\textbf{19.94} & 3.71 \\ \midrule
\multirow{2}{*}{ResNet}        & 18                 & 11.69       & 89.08   & 2.41 &\textbf{21.65} & \textbf{4.69}        \\
                               & 50                 & \textbf{25.56} & \textbf{92.86} & \textbf{4.68} &20.93 & 4.08 \\ \midrule
\multirow{2}{*}{DenseNet}      & 121                & 7.98        & 91.97   & 6.93     &24.20 & \textbf{12.89}     \\
                               & 169                & \textbf{14.15} & \textbf{92.81} & \textbf{10.46} &\textbf{27.15} & 10.60 \\ \bottomrule
\end{tabular}}
\vspace{0.05cm}\caption{Comparison of parameter count vs Adversarial accuracy for five different family of architectures on ImageNet dataset}
\label{paramcount}
\vspace{-0.4cm}
\end{table}

All the results of this comparison are shown in Table \ref{paramcount} and Figure \ref{paramcountfig}. In 4 out of 5 families considered for this study, an increase in parameters increases both clean and adversarial accuracy. The maximum value of the parameter count in these four families in nearly 26 million. This trend of increase in robustness with parameter count is also seen in the fifth family (EfficientNet) but only up to a parameter count of 20 million. Increasing the parameters \textit{alone} beyond 20 million results in a decrease of both clean and adversarial accuracy. This is probably why EfficientNet considers different image sizes for each of the eight networks. After a certain point, increasing the parameters alone will not help improve robustness, and EfficientNet, which has the best adversarial accuracy in the case of ImageNet dataset, conveys this. 

\textit{``In what family of architectures, is the increase in parameter count helping the performance?''} To better understand this, we report PP-HRS in Table \ref{paramcount}. In the case of DenseNAS models developed using MobileNet-V2 search space, an increase in parameters from DenseNAS-A to DenseNAS-Large is improving both clean accuracy and adversarial robustness, which as a result lead to improved PP-HRS. For all the other families, the increased parameter count does not give a significant and sufficient improvement in the PP-HRS and adversarial robustness.

In summary, adversarial robustness can be improved by increasing the number of parameters, but this holds only to an extent. Beyond a certain point (approximately 20-25 million as per our analysis), increasing parameters alone cannot improve adversarial robustness.

\subsection{What makes EfficientNets more robust than other architectures?}
\label{question3}
In comparison to the best performing NAS and hand-crafted architectures in Table \ref{paramcount} (and as discussed in Section~\ref{question1}), the family of  EfficientNet models are significantly better in terms of robustness to adversarial attacks. Among all the architectures compared, EfficientNet-B0 has the highest PP-HRS of $14.90$. In case of PGD, the best performing EfficientNet model, EfficientNet-B4,  outperforms all hand-crafted architectures by atleast 6\%. This is a significant improvement particularly at the scale of ImageNet dataset. Further, for AutoPGD, all EfficientNet models perform better than all hand-crafted architectures (except for DenseNet-169 which is still worse than EfficientNet-B4 and B7).

One significant difference between EffcientNet and existing NAS and hand-crafted models is the scaling factor. Most of the hand-crafted and NAS-based architectures are developed in a micro-style, \ie, a small cell (like the ResNet block or DARTS cell) is developed/searched, and it is stacked to build the full architectures of varying depths and parameter sizes. In the case of EfficientNet, this scaling is done systematically using a compound scaling method using a coefficient ($\phi$) to scale width, depth, and resolution in a principled way\cite{Tanetal20191}. This $\phi$ is specified by the user to control the resources available for scaling the model proportionately in terms of width, depth and resolution. In our analysis, for consistency, we keep the image resolution fixed at $224 \times 224$.

Letting NAS figure out the optimal way to scale a neural network would alleviate the compute required for grid-search (for hyperparameters $\alpha, \beta, \gamma$ in EfficientNet) and makes the complete process of finding an adversarially robust architecture end-to-end. But since Section~\ref{question2} shows that NAS-based architectures are more vulnerable than hand-crafted ones for larger and complex datasets, it is important to better understand the source of this vulnerability to find more effective ways to scale neural networks that are also adverasarially robust. We address this in the next section.

\subsection{Where does the source of adversarial vulnerability lie for NAS?  Is it in the search space or in the way the current methods are performing the search?}\label{question4}

In Section \ref{question2}, we see that NAS-based architectures are more robust than hand-crafted architectures for small-scale datasets and simpler attacks. However, for stronger attacks like PGD, NAS-based architectures are not robust even at the scale of CIFAR-10.
Most of the existing NAS methods perform the search on CIFAR-10 or a subset of ImageNet, and the discovered cell is stacked and trained for other datasets. To understand whether the problem lies in the search space or in the way search is being performed by the existing methods, we performed two simple experiments.

Our first experiment is motivated by \cite{Yang2020NAS}. \cite{Yang2020NAS} shows that a randomly sampled cell in the DARTS search space gives as good a clean accuracy as a searched cell. To test if this fact also holds for the case of adversarial robustness, we sampled random cells from the DARTS search space, stacked and trained them using the standard procedure, and tested their robustness on the CIFAR-10 dataset. Due to the randomness involved, we report the value over four different runs.  Results of this experiment are shown in Table \ref{comparedarts}. We can observed that randomly sampled cells have a better PGD accuracy than the searched architecture. But the variance is very high which shows that relying on randomly sampled architectures for adversarial robustness is not a good idea. This leads us to our second experiment.

For the second experiment, we randomly sample cells from the DARTS search space to build small models (please refer to Figure~\ref{comparison} that illustrates this procedure). After training these models independently, we ensemble the outputs of all these models using a simple linear network. This linear model consists of 2 linear layers with batch normalization and one fully connected classifier layer towards the end that outputs logits based on the number of classes in the dataset. This linear model is just fine-tuned for two epochs. Entire ensemble is treated as one single-network when generating the adversarial examples. To make a fair comparison, we ensure that the ensemble as a whole has the same number of cells as the standard DARTS networks. Since the procedure uses randomly sampled architectures, we run the entire sample-train-ensemble procedure four times and report the mean value in Table \ref{comparedarts}. Due to the randomness involved, this is a computationally expensive procedure. Therefore, we restrict our experiments to CIFAR-10 dataset and DARTS search space.

\begin{table}[h]
\scalebox{0.65}{\begin{tabular}{lccccc}
\toprule
Model            & \# cells & Params (M)  & Clean \%                        & PGD   & AutoPGD       \\ \toprule
DARTS \cite{Darts2018}  & 20        & 3.35        & 97.03                           & 7.09     &6.10    \\
P-DARTS \cite{pdarts}    & 20        & 3.43        & \textbf{97.12} & 9.31   &7.98      \\
PC-DARTS \cite{pcdarts} & 20        & 3.63        & 97.05                           & 9.84    &8.36     \\ \midrule
RANDOM$^\star$     & 20        &    2.73 $\pm$ 0.49         & 95.57 $\pm$ 0.40                           & 14.47 $\pm$ 4.70  & 12.56 $\pm$ 4.16    \\
ENSEMBLE$^\dagger$    & 20        & 2.74 $\pm$ 0.41 & 93.77 $\pm$ 0.39                    & \textbf{21.68 $\pm$ 0.35} & 20.78 $\pm$ 0.39 \\ \bottomrule
\end{tabular}}
\footnotetext[0]\footnotesize{\footnotesize Value reported over four runs.\quad$\star$ Randomly picked architectures from DARTS search-space. Value reported over four runs.}{} 
\footnotetext[0]\footnotesize{\footnotesize$\dagger$ Ensemble of small, randomly picked architectures from DARTS search space.}{}
\caption{\textbf{Ensemble of randomly sampled DARTS cells is significantly more robust than a searched architecture}. Adversarial accuracy comparison of DARTS-based architectures on CIFAR-10.}
\label{comparedarts}
\vspace{-0.3cm}
\end{table}

Surprisingly this simple ensemble of randomly sampled architectures can improve the PGD accuracy of DARTS based models by nearly 12\% and can decrease the variance by $\sim$10\%. Now, this leads to the following interesting conclusions: (1) Learning to build a simple network to combine the outputs of randomly sampled architectures can give clean accuracy with adversarial robustness as an add-on. In this case, we used a simple linear model; replacing this with a searched NAS based architecture can improve the results further. (2) Using NAS to search for an ensemble of architectures can be a potential way to achieve adversarial robustness as an add-on to SoTA clean accuracy. In this case, the NAS objective should be modified to find small models that can complement each other. We plan to explore this in our future work. (3) Both random and ensemble-based topologies are able to provide significantly better adversarial robustness than existing NAS algorithms. This suggests that the search space itself is not the source of vulnerability. Rather, we need better search algorithms, potentially ensemble-based, that can leverage the same search space to build architectures that are inherently more robust even without any explicit adversarial training.

\section{Conclusion}\label{conclusion}

We present a detailed analysis of the adversarial robustness of NAS and hand-crafted models and show how the complex topology of neural networks can be leveraged to achieve adversarial robustness without any form of adversarial training. We also introduce a metric that can be used to calculate the trade-off between clean and adversarial accuracy within and across different families of architectures. Finally, we show that using NAS to find an ensemble of architectures can be one potential way to build robust and reliable models without any form of adversarial training.

\section*{Acknowledgement}
This work has been partly supported by the funding received from DST, Govt of India, through the Data Science cluster of the ICPS program (DST/ICPS/CLUSTER/Data Science/2018/General).

{\small
\bibliographystyle{ieee_fullname}
\bibliography{egbib}
}
 
\clearpage
\appendix

\section{Supplementary Material}
\vspace{-5pt}
In this supplementary section, we discuss the following details, which could not be included in the paper owing to space constraints.
\vspace{-10pt}
\begin{itemize}
 \itemsep -0.4em
 \item Details of the randomly sampled architectures used in Section \ref{question4}.
    \item A quantitative and qualitative comparison of searched, randomly sampled architectures used in Section \ref{question4}.
    \item Comparison of NAS and hand-crafted architectures for black-box attacks on CIFAR-10 dataset.
\end{itemize}

\vspace{-0.4cm}

\subsection{Details of Randomly sampled architectures from DARTS search space} \label{ap1}

In Section \ref{question4}, we report results over four random runs, details of each of these runs are shown in Table \ref{details-arch}. `\# Networks' denotes the number of sub-networks in a given ensemble, and `\# Cells' denote the number of cells in each sub-networks. `\# Epochs' denotes the number of epochs each sub-network is trained for. 

\vspace{-0.4em}
\begin{table}[h]
\scalebox{0.85}{\begin{tabular}{@{}cccc@{}}
\toprule
Run \# & \begin{tabular}[c]{@{}c@{}}\# Networks\end{tabular} & \begin{tabular}[c]{@{}c@{}}\# Cells \end{tabular} & \begin{tabular}[c]{@{}c@{}}\# Epochs\end{tabular} \\ \midrule
1      & 3                                                         & \{12, 6, 2\}                                                           & \{360, 180, 40\}                                                                \\
2      & 5                                                         & \{4, 4, 4, 4, 4\}                                                      & \{120, 120, 120, 120, 120\}                                                     \\
3      & 5                                                         & \{6, 5, 4, 3, 2\}                                                          & \{180, 150, 120, 90, 60\}                                                       \\
4      & 3                                                         & \{16, 2, 2\}                                                               & \{480, 60, 60\}                                                                 \\ \bottomrule
\end{tabular}}
\caption{Details of sub-networks in each ensemble across four runs}
\label{details-arch}
\end{table}

\vspace{-0.65cm}

\subsection{Comparison of Searched and Randomly sampled architectures}
\vspace{-0.2cm}
In this section, we make a qualitative and quantitative comparison between the randomly sampled architectures (used in Section \ref{question4}) with standard SoTA DARTS based architectures like DARTS \cite{Darts2018}, P-DARTS \cite{pdarts}, and PC-DARTS \cite{pcdarts}. For this study, we randomly choose 3 of the 16 randomly sampled sub-networks shown in Table \ref{details-arch}.

\begin{table}[h]
\scalebox{0.75}{\begin{tabular}{@{}lccccc|c@{}}
\toprule
         & \begin{tabular}[c]{@{}c@{}}Max\\ Pool\end{tabular} & \begin{tabular}[c]{@{}c@{}}Avg\\ Pool\end{tabular} & \begin{tabular}[c]{@{}c@{}}Skip\\ connection\end{tabular} & \begin{tabular}[c]{@{}c@{}}Separable\\ Conv\end{tabular} & \begin{tabular}[c]{@{}c@{}}Dilated\\ Conv\end{tabular} & \begin{tabular}[c]{@{}c@{}}\# unique\\ operations\end{tabular} \\ \midrule
DARTS    & 0                                                  & 0                                                  & 2                                                         & 5                                                        & 1                                                      & 3                                                              \\
P-DARTS  & 0                                                  & 0                                                  & 2                                                         & 4                                                        & 2                                                      & 3                                                              \\
PC-DARTS & 0                                                  & 1                                                  & 1                                                         & 4                                                        & 2                                                      & 4                                                              \\ \midrule
R1       & 1                                                  & 1                                                  & 2                                                         & 1                                                        & 3                                                      & \textbf{5}                                                     \\
R2       & 1                                                  & 1                                                  & 3                                                         & 2                                                        & 1                                                      & \textbf{5}                                                     \\
R3       & 1                                                  & 1                                                  & 1                                                         & 2                                                        & 3                                                      & \textbf{5}                                                     \\ \bottomrule
\end{tabular}}
\caption{Comparison on usage of different operations in \textbf{normal} cell of micros from DARTS search space; R1, R2, R3 denote three randomly sampled micros.}
\label{normal-table}
\vspace{-0.4cm}
\end{table}

DARTS search space consists of 5 operations: Max Pooling, Average Pooling, Skip Connections, Separable and Dilated convolutions. We report the number of times these operations are used in the normal and reduce cells of different DARTS architectures/micros in Tables \ref{normal-table} and \ref{reduce-table}. When compared with searched cells, randomly sampled ones, in general, have more number of unique operations. In a searched micro, the maximum number of unique operations for the normal, reduce cells in three, four, respectively. For a randomly sampled architecture, the count is five for both normal and reduce cells. Searched cells have many occurrences of a single operation (Separable convolution), which is not the case in randomly sampled architectures. We hypothesize that the presence of diverse set operations is a plausible for improved adversarial accuracy of randomly sampled DARTS architectures. While we only show 3-randomly chosen sub-networks in Tables \ref{normal-table}, \ref{reduce-table}, we observe similar inferences with other random choices for sub-networks too.  A qualitative comparison of these operations is shown in Figures \ref{normal-cell-fig} and \ref{reduction-cell-fig}.

\begin{table}[h]
\scalebox{0.75}{\begin{tabular}{llcccc|c}
\toprule
         & \begin{tabular}[c]{@{}l@{}}Max\\ Pool\end{tabular} & \begin{tabular}[c]{@{}c@{}}Avg\\ Pool\end{tabular} & \begin{tabular}[c]{@{}c@{}}Skip\\ connection\end{tabular} & \begin{tabular}[c]{@{}c@{}}Separable\\ Conv\end{tabular} & \begin{tabular}[c]{@{}c@{}}Dilated\\ Conv\end{tabular} & \begin{tabular}[c]{@{}c@{}}\# unique \\ operations\end{tabular} \\ \toprule
DARTS    & 5                                                  & 0                                                  & 3                                                         & 0                                                        & 0                                                      & 2                                                               \\
P-DARTS  & 1                                                  & 1                                                  & 0                                                         & 2                                                        & 4                                                      & 3                                                               \\
PC-DARTS & 1                                                  & 0                                                  & 0                                                         & 7                                                        & 0                                                      & 2                                                               \\ \midrule
R1       & 1                                                  & 1                                                  & 1                                                         & 2                                                        & 3                                                      & \textbf{5}                                                      \\
R2       & 1                                                  & 1                                                  & 2                                                         & 3                                                        & 1                                                      & \textbf{5}                                                      \\
R3       & 1                                                  & 1                                                  & 2                                                         & 3                                                        & 1                                                      & \textbf{5}                                                      \\ \bottomrule
\end{tabular}}
\caption{Comparison on usage of different operations in \textbf{reduce} cell of micros from DARTS search space; R1, R2, R3 denote three randomly sampled micros;}
\label{reduce-table}
\vspace{-0.4cm}
\end{table}

\subsection{More Results: Adversarial Accuracy on Black-box Attacks}
Table \ref{c10-bb} shows the robustness of different architectures for black-box attacks on CIFAR-10. In a black-box setting, the robustness of a model is tested on adversarial examples generated using a source model. For the source model, we use two variants of hand-crafted models and two variants of NAS models.  Since we use no adversarial training in our experiments, the accuracy numbers are relatively lower. In general, the average adversarial accuracy on hand-crafted architectures is higher than NAS methods, having complex operations in the topology (DenseNet, ProxylessNAS) or a large number of parameters (VGG-16) provide the best adversarial accuracy in a black-box setting. Either way, our hypothesis that the complexity of the architecture (in terms of operations or parameters) plays a major role is corroborated in this study too. Understanding specific details of the nature and origin of such complexity in topology may be an interesting direction of future work.

\begin{table}[h]

\scalebox{0.8}{\begin{tabular}{@{}lcc|cc@{}}
\toprule
Target $\downarrow$  Source $\rightarrow$ & ResNet-18      & DesneNet-169   & DARTS          & NSGA Net       \\ \midrule
ResNet-18      & 9.59           & 9.51           & 9.59           & 9.60           \\
ResNet-50      & 10.96          & 10.98          & 10.97          & 10.92          \\
DenseNet-121   & 11.69          & 11.67          & 11.66          & 11.66          \\
DenseNet-169   & 9.74           & 9.71           & 9.74           & 9.71           \\
VGG16 BN       & \textbf{13.96} & \textbf{13.64} & \textbf{13.85} & \textbf{13.94} \\ \midrule
DARTS          & 10.05          & 10.04          & 10.04          & 10.05          \\
PDARTS         & 9.95           & 9.98           & 10.07          & 10.11          \\
NSGA Net       & 9.85           & 9.86           & 9.90           & 9.88           \\
Proxyless-NAS  & 11.97          & 11.89          & 11.95          & 11.97          \\
PC-DARTS       & 7.96           & 8.02           & 8.1            & 7.94           \\ \bottomrule
\end{tabular}}
\caption{Comparison of adversarial accuracy for different Black-Box attacks on CIFAR-10 dataset}
\label{c10-bb}
\end{table}

\begin{figure*}
\begin{center}
\includegraphics[width=18cm, height=9.8cm]{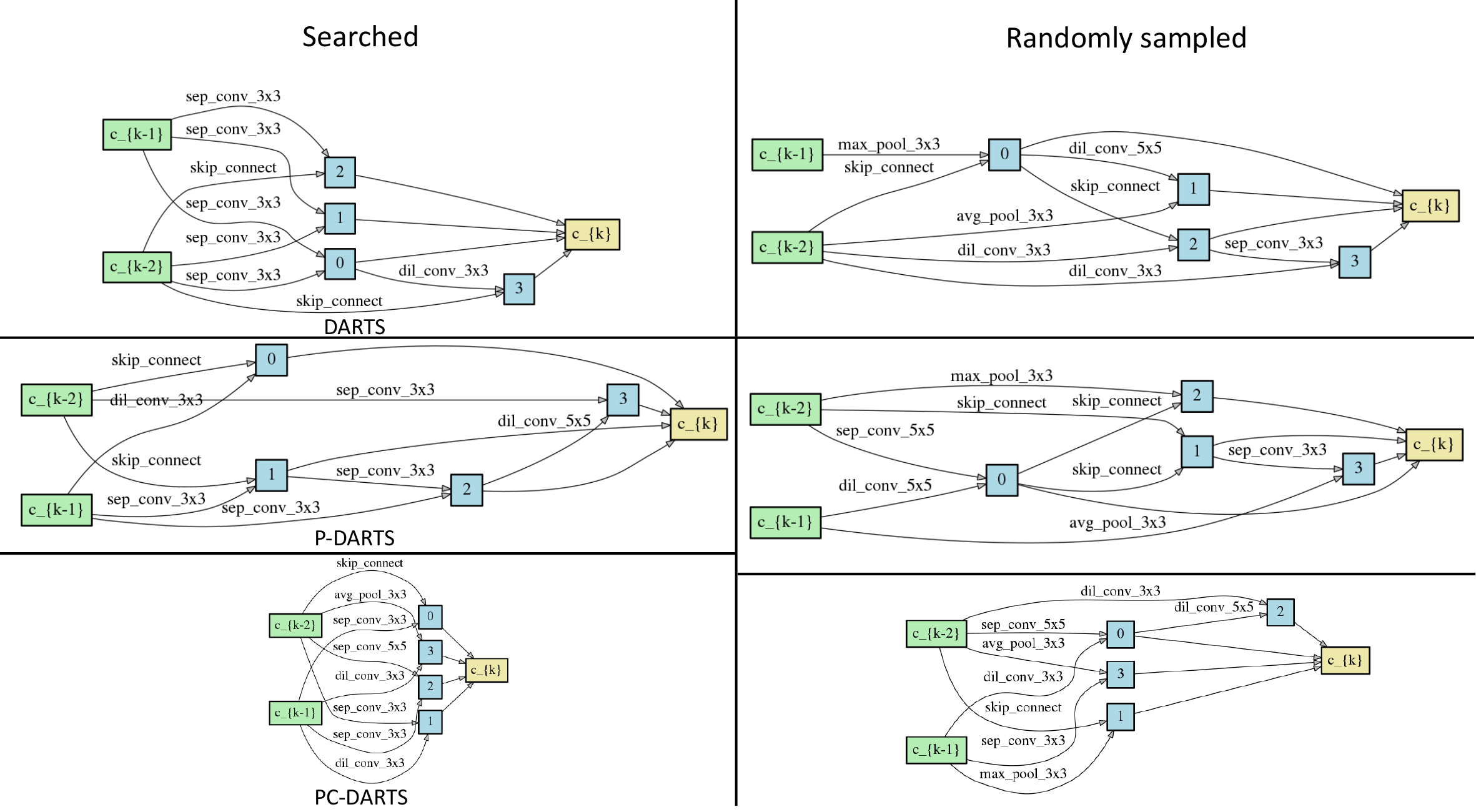}
\end{center}
   \caption{Qualitative comparison of different operations in \textbf{normal} cells of searched, randomly sampled micros from DARTS search space; \textit{Left:} Searched micros, \textit{Right:} Randomly sampled micros; Randomly sampled ones, in general, have more unique operations than searched ones} \label{normal-cell-fig}
\vspace{-0.5cm}
\label{normal-cell-fig}
\end{figure*}

\begin{figure*}
\begin{center}
\includegraphics[width=18cm, height=9.8cm]{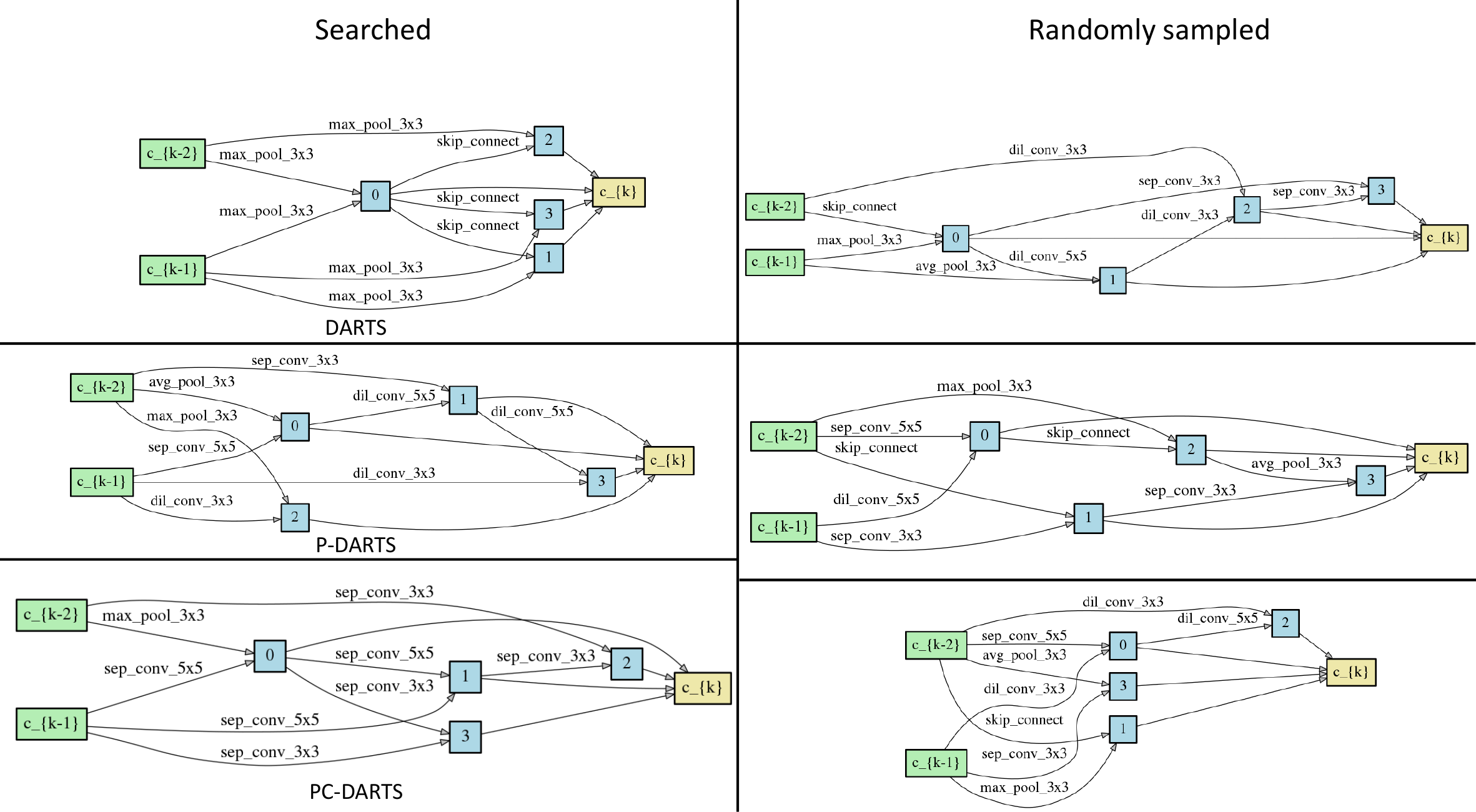}
\end{center}
   \caption{Qualitative comparison of different operations in \textbf{reduce} cells of searched, randomly sampled micros from DARTS search space; \textit{Left:} Searched micros, \textit{Right:} Randomly sampled micros; Randomly sampled ones, in general, have more unique operations than searched ones} \label{reduction-cell-fig}
\vspace{-0.5cm}
\label{reduction-cell-fig}
\end{figure*}

\end{document}